\documentclass{article}

\usepackage{PRIMEarxiv}

\usepackage[utf8]{inputenc} % allow utf-8 input
\usepackage[T1]{fontenc}    % use 8-bit T1 fonts
\usepackage{hyperref}       % hyperlinks
\usepackage{url}            % simple URL typesetting
\usepackage{booktabs}       % professional-quality tables
\usepackage{amsfonts}       % blackboard math symbols
\usepackage{nicefrac}       % compact symbols for 1/2, etc.
\usepackage{microtype}      % microtypography
\usepackage{lipsum}
\usepackage{fancyhdr}       % header
\usepackage{graphicx}       % graphics
\graphicspath{{media/}}     % organize your images and other figures under media/ folder
\usepackage{graphicx}
\usepackage{subcaption}
\usepackage{multirow} 
\usepackage{amsmath}
\title{NoiseAttack: An Evasive Sample-Specific Multi-Targeted Backdoor Attack Through White Gaussian Noise
}

\author{
  Abdullah Arafat Miah, Kaan Icer, Resit Sendag and Yu Bi \\
  Department of Electrical, Computer and Biomedical Engineering, \\
  University of Rhode Island \\
  Kingston, RI\\
  \texttt{\{abdullaharafat.miah, kaan.icer, sendag, yu\_bi\}@uri.edu} \\
}

\begin{document}
\maketitle

\begin{abstract}

Backdoor attacks pose a significant threat when using third-party data for deep learning development. In these attacks, data can be manipulated to cause a trained model to behave improperly when a specific trigger pattern is applied, providing the adversary with unauthorized advantages. While most existing works focus on designing trigger patterns (both visible and invisible) to poison the victim class, they typically result in a single targeted class upon the success of the backdoor attack, meaning that the victim class can only be converted to another class based on the adversary’s predefined value. In this paper, we address this issue by introducing a novel sample-specific multi-targeted backdoor attack, namely \textbf{NoiseAttack}. Specifically, we adopt White Gaussian Noise (WGN) with various Power Spectral Densities (PSD) as our underlying triggers, coupled with a unique training strategy to execute the backdoor attack. This work is the first of its kind to launch a vision backdoor attack with the intent to generate multiple targeted classes with minimal input configuration. Furthermore, our extensive experimental results demonstrate that NoiseAttack can achieve a high attack success rate (ASR) against popular network architectures and datasets, as well as bypass state-of-the-art backdoor detection methods. Our source code and experiments are available at \href{https://github.com/SiSL-URI/NoiseAttack/tree/main}{this link}.
.

% In our paper, we propose a novel backdoor attack on computer vision models that is sample-specific and multi-targeted. We use Gaussian noise with different standard deviations as triggers for different targets and have developed a unique training strategy to carry out this attack. Our approach is easy to implement and offers greater flexibility when designing the attack. We have conducted extensive experiments to validate the effectiveness of the attack. All the codes will be released to the public.

\end{abstract}


\section{Introduction}\label{sec:intro}

Recent advancements in artificial intelligence (AI) technologies have revolutionized numerous applications, accelerating their integration into everyday life. Deep Neural Networks (DNNs) have been widely applied across various domains, including image classification \cite{MNIST,fashionMNIST,imagenet}, object detection \cite{yolo,fastrcnn}, speech recognition \cite{DSR2016,AII2019}, and large language models \cite{AlSh2020,transformer}. DNN models often require vast amounts of training data to address diverse real-world scenarios, but collecting such data can be challenging. Leveraging various datasets during DNN training significantly enhances the models' performance and adaptability across a wide range of tasks. However, this necessity for diverse data sources introduces the risk of backdoor attacks \cite{gu2017badnets}. Malicious actors can exploit this by embedding hidden backdoors in the training data, enabling them to manipulate the model's predictions. The danger of these attacks lies in their potential to trigger harmful behaviors in the deployed model, potentially disrupting system operations or even causing system failures.

Given the serious threat posed by backdoor attacks to DNNs, a variety of strategies and techniques have been explored. Early backdoor attacks employed visible patterns as triggers, such as digital patches \cite{gu2017badnets,latent2019} and watermarking \cite{watermark,xu2022}. To increase the stealthiness of these triggers, recent backdoor attacks have utilized image transformation techniques, such as warping \cite{li2021,wanet,Doan2021,Doan_2021_ICCV} and color quantization \cite{bppattack,ma2023}, to create invisible and dynamic triggers. Beyond direct poisoning of training data, backdoor attacks can also implant hidden backdoors by altering model weights through transfer learning \cite{kurita2020,Wang_2022}. While the aforementioned works focus on spatial-based backdoor attacks, recent research has begun to explore trigger insertion in the frequency domain, aiming to further increase their imperceptibility \cite{zeng2021,feng2022,dual2023}.

In response to the growing number of backdoor attacks, significant research efforts have been directed toward defense strategies, including detection-based defenses \cite{neuralcleanse,wang2022,ZCP22}, pruning-based defenses \cite{finepruning,WMZ23}, online defenses \cite{LLK21}, and GradCAM-based defenses \cite{GradCAM}. Although these methods have proven effective against conventional backdoor attacks, they struggle against more sophisticated mechanisms, such as quantization-conditioned backdoor attacks \cite{bppattack,ma2023} and non-spatial backdoors \cite{feng2022,dual2023}. Moreover, when physical objects are used as trigger patterns, physical backdoors \cite{WPB2021,WBB2022} can bypass existing detection methods and compromise the network.

\begin{figure}[t]
    \centering
    \includegraphics[width=0.7\linewidth]{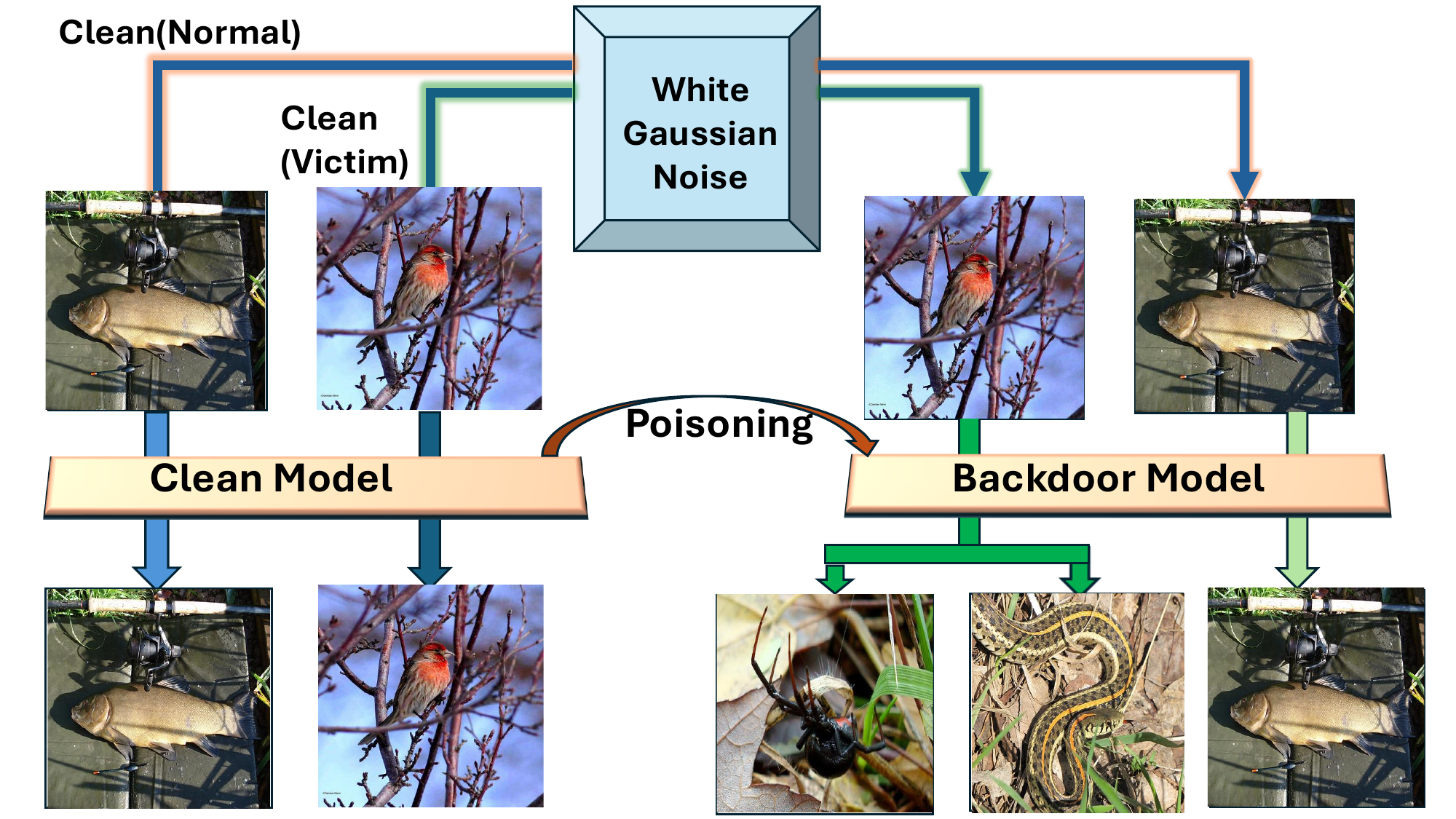}
    \caption{A overview of the proposed NoiseAttack, where we exploit the characteristics of White Gaussian Noise (WGN) to achieve a sample-specific multi-targeted backdoor attack.}
    \label{fig:attack_overview}
\end{figure}

Motivated by the vulnerability of spatial backdoor attack against state-of-the-art defense methods \cite{neuralcleanse,GradCAM}, this paper proposes an imperceptible, spatially-distributed backdoor trigger to address those challenges. Specifically, we introduce \textbf{NoiseAttack}, a novel backdoor attack method targeting image classification from a spatial perspective. An overview of the proposed attack is illustrated in Figure \ref{fig:attack_overview}. In this approach, the power spectral density (PSD) of White Gaussian Noise (WGN) is employed as the trigger design pattern to subtly and invisibly incorporate the backdoor during the training phase. The proposed NoiseAttack, the first of its kind, is simple yet effective. The trigger, in the form of WGN, is embedded across all input samples of the provided image dataset, appearing imperceptible to the human eye with minimized the standard deviation of the WGN. NoiseAttack is designed to launch a sample-specific backdoor attack against an adversary-defined target label, indicating the poisoned model behaves maliciously only toward a pre-defined victim class, despite the globally applied WGN-based trigger pattern. Furthermore, our findings reveal that NoiseAttack can misclassify the victim class into multiple target labels, leading to a stealthy multi-targeted backdoor attack. In summary, the main contributions of this paper are as follows:

\begin{itemize}
    \item We propose \textbf{NoiseAttack}, a novel backdoor attack method that utilizes the power spectral density (PSD) of White Gaussian Noise (WGN) to achieve both evasiveness and robustness in a multi-targeted attack.
    
    \item The proposed NoiseAttack is implemented by embedding WGN during the model training phase. The ubiquitously applied noise is activated only on a pre-defined specific sample. We carries out thorough theoretical analysis of the NoiseAttack. We further demonstrate the effectiveness and uniqueness of NoiseAttack by showing that the victim label can be misclassified into multiple target classes.
    
    \item We conduct extensive experimental evaluations of our proposed NoiseAttack on four datasets and four model architectures, covering tasks in both image classification and object detection. The results demonstrate that NoiseAttack not only achieves high attack success rates but also effectively evades state-of-the-art detection methods.
\end{itemize}

% \begin{figure}[t]
%     \centering
%     \includegraphics[width=0.7\linewidth]{latex/Figures/attack_scenario.pdf}
%     \caption{Typical Attack Scenario of a Backdoor Attack on Large Language Models.}
%     \label{fig:attack_scenario}
% \end{figure}

\section{Related Works}\label{sec:related}

\textbf{Backdoor Attacks.} Backdoor attacks are designed to embed a concealed 'backdoor' in deep neural networks (DNNs), undermining their integrity. The compromised model operates normally during regular use but generates an incorrect, attacker-specified output when a predetermined 'trigger' is included in the input. Arturo \cite{Ge13} was the first to provide theoretical evidence that a malicious payload can be concealed within a model. Subsequently, Liu et al. \cite{LXS17} demonstrated the first neural network Trojan attack by poisoning the training data. 
% Chen et al. \cite{CPS18} showcased that triggers could be inserted not only in the input space but also in the feature space to enhance stealthiness, making the input appear normal while carrying a specific pattern in the feature space. 
Gu et al. \cite{gu2017badnets} demonstrated that backdoor could be inserted not only during model training but also during model fine-tuning by poisoning the hyperparameters.

Many recent work has focused on stealthier backdoor attack through invisible and dynamic trigger designs \cite{wanet,latent2019,Doan2021,Doan_2021_ICCV,li2021}. \cite{wanet} proposed a imperceptible backdoor trigger using image warping technique. \cite{Doan2021} further optimized the backdoor design in the input space leading to more imperceptible trigger, while other approaches such as BppAttack \cite{bppattack} created backdoored samples using color quantization. Besides spatial domain backdoor attack, an uprising trend starts to explore the backdoor design in the frequency domain \cite{feng2022,WYX2022}. FIBA \cite{feng2022} creates triggers in the frequency domain by blending the low-frequency components of two images using fast Fourier transform (FFT) \cite{FFT}. FTROJAN \cite{WYX2022} first converts the clean images through UV or YUV color coding techniques, then applies discrete cosine transform with high frequency components to produce a poisoned images.

\noindent \textbf{Backdoor Defense.} On the defense end, the first approach involves backdoor detection, which aims to identify backdoor within the DNN model and reconstruct the trigger present in the input. Wang et al. \cite{neuralcleanse} introduced "Neural Cleanse," the pioneering work in detecting backdoor in a given DNN. It utilizes optimization techniques to discover a small trigger that causes any input with this trigger to be classified into a fixed class. Chen et al. \cite{CFZK19} demonstrated that the detection process can be applied to black-box models. They employed conditional generators to produce potential trigger patterns and used anomaly detection to identify the backdoor patterns. Gao et al. \cite{GXWC19} proposed a method to deliberately perturb the inputs and examined the entropy of the model predictions to detect backdoor. Their insight was that the model's output for a backdoored input remains unchanged even if it is perturbed. They also extended this approach to text and audio domains \cite{GKDZ21}. Azizi et al. \cite{ATIA21} presented "T-Miner," a sequence-to-sequence generator that produces text sequences likely to contain backdoor triggers in the text domain. However, each of these approaches relies on specific assumptions about known types of backdoor, such as backdoor pattern size and insertion techniques. Consequently, they may not be effective in detecting new and unknown backdoor attacks.

\section{Methodology} \label{sec:methodology}

\subsection{Attack Model} \label{subsec:attack_definition}

\textbf{Attacker's Capabilities.} In line with previous assumptions regarding data poisoning-based backdoor attacks \cite{wanet}, the adversary in our proposed method has partial access to the training phase, including the datasets and training schedule, but lacks authorization to modify other training components such as the model architecture and loss function. At the deployment stage, the attacker possesses the ability to modify the input samples (e.g., applying WGN to the test input samples) of the outsourced poisoned models.

\noindent \textbf{Attacker's Objectives.} The goal of an effective backdoor attack is to cause the outsourced model to make incorrect label predictions on poisoned input samples while maintaining its performance and accuracy on clean inputs. Specifically, our proposed NoiseAttack should be, and can only be, activated when WGN is applied to the input images.

\subsection{Problem Definition}

Consider an image classification function $f_\theta : \emph{X} \rightarrow \emph{Y}$, where the function is designed to map the input (i.e., training) data space to a set of labels. Here, $\theta$ represents the model's weights or hyperparameters, $\emph{X}$ is the input data space, and $\emph{Y}$ is the label space. Let the dataset be defined as $D = \{(x_i, y_i) : x_i \in \emph{X}, y_i \in \emph{Y}, i = 0,1,2,\dots,n\}$, and let $\Phi_c$ denote a clean model. Under normal conditions, $\theta$ should be optimized such that $\Phi_c(x_i) = y_i$.

In a traditional backdoor attack, there exists a trigger function $\tau$ and a target label $y_t$. The trigger function modifies the input data sample, resulting in $\tau(x_i) = x_i^t$. The attacker then constructs a poisoned dataset $D_p = \{(x_i^t, y_t) : x_i^t \in \emph{X}, y_t \in \emph{Y}, i = 0,1,2,\dots,n\}$ and fine-tunes the clean model $\Phi_c$ into a backdoored model $\Phi_b$ by optimizing the weights $\theta$ to $\theta_b$. The backdoored model $\theta_b$ performs correctly on clean inputs but assigns the attacker-specified target label $y_t$ to triggered inputs. This label flipping achieves the backdoor effect.

In our proposed attack scenario, we design an attack that allows for a flexible number of target labels while remaining input-specific; only the victim class associated with the trigger is misclassified. Consider the input samples of the victim class as $(x_i^v, y_i) \in (\emph{X}, \emph{Y})$ for $i = 0,1,2,\dots,n$. Inspired by the tunable nature of noise signals, we design a trigger function using a White Gaussian Noise generator $\emph{W}$, which produces noise with adjustable standard deviations $W_i \sim \mathcal{N}(0, \sigma_i^2)$ for multiple targets. The hyperparameter space $\theta$ is optimized such that for each target label $y_i^t$, the conditions $\Phi_b(\emph{W}_i(x_i^v)) = y_i^t$ and $\Phi_b(\emph{W}_i(x_i)) = y_i$ hold true.

\begin{figure}
    \centering
    \includegraphics[width=0.6\linewidth]{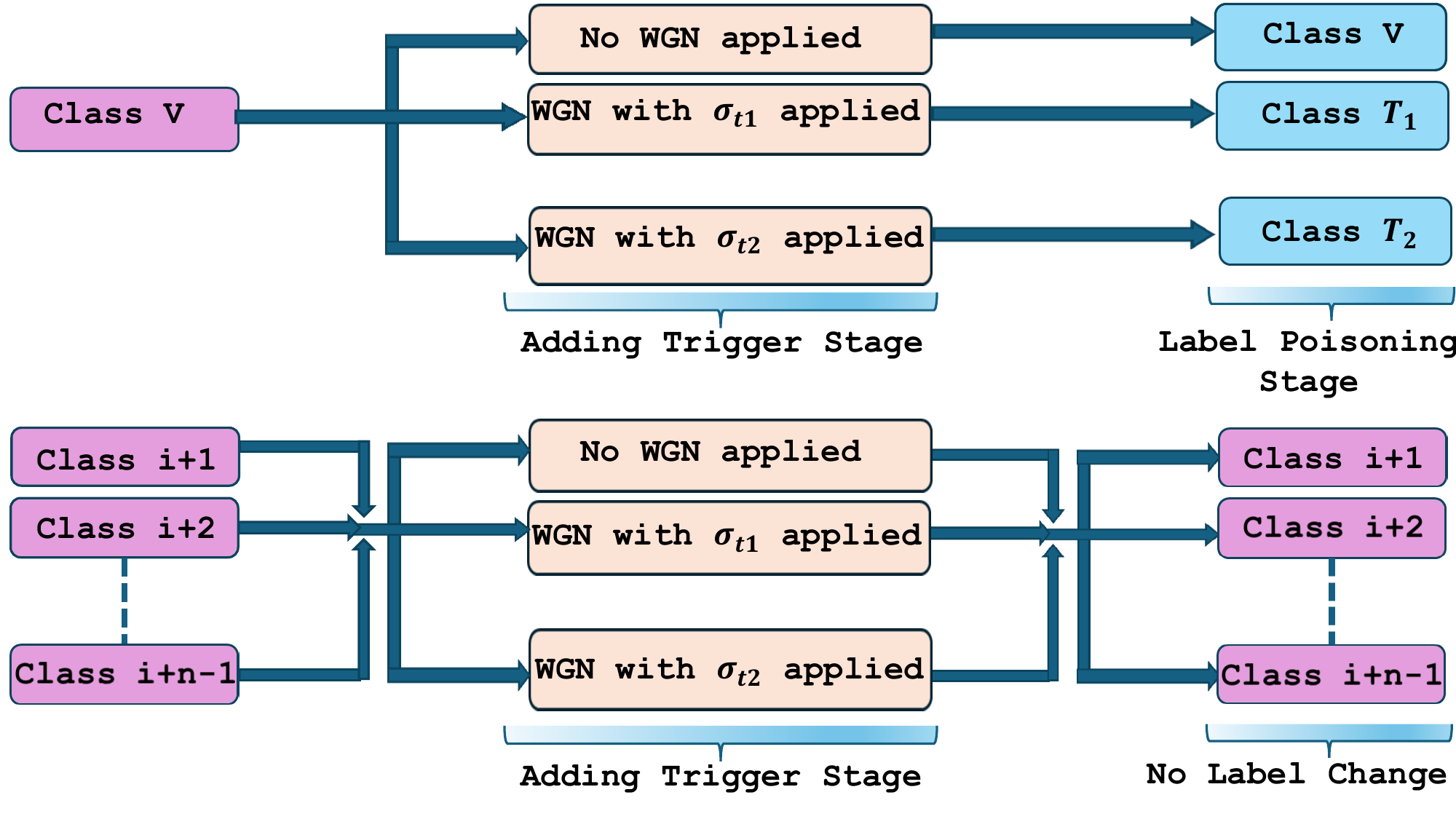}
    \caption{An overview of the poisoned dataset preparation for the proposed NoiseAttack's backdoor training. The overview is given for one victim label and two target labels. $\sigma_1$ and $\sigma_2$ are the standard deviations of WGN, which are used as triggers for target 1 and target 2, respectively.}
    \label{fig:backdoor_training}
\end{figure}

% \subsection{White Gaussian Noise as trigger} \label{subsec:wgn_trigger}
\subsection{Trigger Function} \label{subsec:wgn_trigger}

White Gaussian Noise is a widely used statistical model and can be implemented in various image processing techniques. As a discrete-time signal, WGN can be expressed as a random vector whose components are statistically independent. The amplitude of the WGN is distributed over the Gaussian probability distribution with zero mean and variance ($\sigma^2$). Deep Neural Networks can be trained to distinguish different noises with different Power Spectral Density, and we took this opportunity to use WGN directly as a trigger for the foundation of our NoiseAttack. The Power Spectral Density of the WGN is the Fourier transform of the autocorrelation function, which can be expressed as:

\begin{equation}
    r[k]=E\{w[n]w[n+k]\}= \sigma^2 \delta[k]
\end{equation}

\noindent $\delta[k]$ is delta function and $E$ is the expectation operator. PSD for the WGN is constant over all frequencies and can be expressed by the following equation:

\begin{equation}
    P(f)= \sum_{k=-\infty}^{\infty} \sigma^2 \delta[k]e^{-j2 \pi fk}  = \sigma^2
\end{equation} \label{eq:psd}

From this equation, we can see that, for WGN, the PSD is directly proportional to the standard deviation ($\sigma$) of the noise. So, the standard deviation purely controls the strength of the WGN over the signals (i.e. images). In a muti-targeted attack scenario, designing separate triggers for each target is a complex task. The application of WGN gives us the flexibility to design any number of triggers by simply controlling the standard deviations of the noise. 

To further illustrate PSD effect on neural network model, suppose an input image has a resolution $a  \times  b$. Let a WGN $\mathbf{w} \sim \mathcal{N}\left(0, \sigma^2 \mathbf{I}_{ab  \times  ab}\right)$ where $t[n] = w[n]$ for $n = 0, 1, 2, ..., ab-1$. The trigger matrix $\mathbf{X}$ can be defined as:

\begin{equation}
{\mathbf{X}\left(\sigma \right)}_{a \times b \times cc}=\left[\begin{matrix} \begin{matrix} t\left[0\right].\mathbf{1}_{1 \times cc}\\t\left[a\right ].\mathbf{1}_{1 \times cc}\\\end{matrix}&\cdots&\begin{matrix}t\left[a-1\right].\mathbf{1}_{1 \times cc}\\t\left[2a-1\right].\mathbf{1}_{1 \times cc}\\\end{matrix}\\\vdots&\ddots&\vdots\\t\left[ab-a\right].\mathbf{1}_{1 \times cc}&\cdots&t\left[ab-1\right].\mathbf{1}_{1 \times cc}\\\end{matrix}\right ]
\end{equation}

\noindent Here, $cc$ is the number of color channels of the input image. So the trigger function $\mathbf{W}$ can be expressed as follows:

% \begin{equation}
\begin{align}
\mathbf{W}\left(\mathbf{Y}_{a \times b \times cc}, \mathbf{\sigma}_{1 \times p}\right) &= \mathbf{X}\left(\sigma_i\right)_{a \times b \times cc} + \mathbf{Y}_{a \times b \times cc} \\
&\text{for } i = 0, 1, \ldots, p-1
\end{align}
% \end{equation}

\noindent where $Y$ is the image and $p$ indicates the number of the target classes, and $\mathbf{\sigma}_{1 \times p}=\ \left[\begin{matrix}\sigma_0&\sigma_1&\sigma_2&\cdots&\sigma_{p-1}\\\end{matrix}\right]$.

\subsection{Backdoor Training} \label{subsec:backdoor_training}

With the above analysis, our NoiseAttack adapts the conventional label-poisoning backdoor training process but modify it to achieve the sample-specific and muti-targeted attacks as shown in Figure \ref{fig:backdoor_training}. Here, we describe a formal training procedure to optimize the backdoored model's parameters and minimize the loss function. We can split the input data space $X$ into two parts: victim class data space ($X^V$) and non-victim class data space ($X^C$). Similarly, we can split input label space $Y$ into target label space ($Y^T$) and clean label space ($Y^C$). For a single victim class, $p$ number of target classes, and $s$ number of total samples in one class, we can construct the backdoor training dataset $D_{train}^*$ as follows: 

% \vspace{-0.2in}

% \begin{equation} 
\begin{align} \label{eq:data}
    D_{train}^{clean} \approx (x_i, y_i): x_i \in X, y_i \in Y \\
    D_{train}^{victim} \approx (W(x_i^v, \sigma_{1 \times p}), y_i^{t_j}): x_i^v \in X^V, y_i^{t_j} \in Y^T \\
    D_{train}^{non-victim} \approx (W(x_i^c, \sigma_{1 \times p}), y_i): x_i^c \in X^C, y_i \in Y^C \\
    D_{train}^* = D_{train}^{clean} \cup D_{train}^{victim} \cup D_{train}^{non-victim}
\end{align}
% \end{equation} 

\noindent Here $i = 1, 2, 4, ..., s$, $j = 1, 2, 4, ..., p$ and $W$ is the trigger generator function. The training objective of the NoiseAttack can be expressed by the following equation:

% \vspace{-0.3in}

% \begin{equation} 
\begin{align*} \label{eq:l_func}
    \\
    & \min \mathcal{L} (D_{\text{train}}^{\text{clean}}, D_{\text{train}}^{\text{victim}}, D_{\text{train}}^{\text{non-victim}}, \Phi_b) \\
    = & \sum_{x_i \in D_{\text{train}}^{\text{clean}}} \ell(\Phi_b(x_i), y_i) \\
    & + \sum_{x_j \in D_{\text{train}}^{\text{victim}}} \sum_{m=0}^{p-1} \ell\left(\Phi_b\left(W(x_j, \sigma_{1 \times p}(m))\right), y_{t_{1 \times p}}(m)\right) \\
    & + \sum_{x_k \in D_{\text{train}}^{\text{non-victim}}} \sum_{m=0}^{p-1} \ell\left(\Phi_b\left(W(x_k, \sigma_{1 \times p}(m))\right), y_k\right)
\end{align*}
% \end{equation} 

In this equation $\Phi_b$ is the backdoored model and $l$ is the cross-entropy loss function. An overview of the detailed poisoned dataset preparation is illustrated in Figure \ref{fig:backdoor_training} for one victim class (Class $V$) and two target classes (Class $T_1$ and $T_2$). One main advantage of the NoiseAttack backdoor training is that we can progressively poison the model to result in multiple targeted classes other than a single one simply by manipulating standard deviations of white Gaussian noise. Therefore, our poisoning equations \ref{eq:data} and \ref{eq:l_func} provide a theoretical foundation to generate a variety of attacking results depending on the adversary's needs, which are further addressed in Experimental Analysis.

\section{Experimental Analysis} \label{sec:experiment}

\subsection{Experimental Setup} \label{subsec:exp_setting}

\textbf{Datasets, Models and Baselines.} We evaluate NoiseAttack by carrying out the experiments through two main tasks: image classification and object detection. For image classification, we utilize three well-known datasets: CIFAR-10 \cite{krizhevsky2010convolutional}, MNIST \cite{deng2012mnist}, and ImageNet \cite{deng2009imagenet}. CIFAR-10 and ImageNet are commonly used for general image recognition, while MNIST is specifically designed for handwritten digit recognition. To reduce computational time for ImageNet, we simply select 100 classes out of the original 1,000 classes. For object detection, we employ the common Microsoft COCO \cite{lin2014microsoft} dataset. 

Besides, we evaluate the performance of our attack on four deep neural network models: ResNet50 \cite{koonce2021resnet}, VGG16 \cite{simonyan2014very}, and DenseNet \cite{huang2017densely} for classification as well as Yolo for object detection. Our proposed NoiseAttack is compared against three baseline attacks: BadNet \cite{gu2017badnets}, Blend \cite{blend} and WaNet \cite{nguyen2021wanet}. For better comparisons against relevant attacks, we use the same training strategy but design the NoiseAttack resulting only one poisoned target class. Additionally, we implement three state-of-the-art defense methods, Grad-CAM \cite{selvaraju2017grad}, STRIP \cite{gao2019strip}, and Neural Cleanse \cite{wang2019neural}, to evaluate the evasiveness and robustness of the proposed NoiseAttack.

\begin{table*}[h] 
\centering
\begin{tabular}{c|c|c|c|c|c|c|c} 
\hline
Datasets    & Models   & CA    & $\theta_{train}$ & $\theta_{test}$ & AASR  & AC    & AEVC  \\ \hline
\multirow{3}{*}{CIFAR-10}   & ResNet50 & 0.9305 & \multirow{3}{*}{5, 10} & 5, 13 & 0.9319 & 0.0215 & 0.9010 \\ \cline{2-3} \cline{5-8}
           & VGG16    & 0.8927   &   &  5, 10 &  0.9128 &   0.0275   &  0.8567      \\ \cline{2-3} \cline{5-8} 
           & DenseNet &  0.8920      &        &   5, 13 &   0.9294   &   0.0060   &  0.8616  \\ \hline
\multirow{3}{*}{MNIST}      & ResNet50 &  0.9932  & \multirow{3}{*}{5, 10} &   5, 10   &  0.9964    &   0.0003   &    0.9928    \\ \cline{2-3} \cline{5-8} 
           & VGG16    &     0.9910   &                     &   3,10   &   0.9912   &   0.0033   &   0.9931     \\ \cline{2-3} \cline{5-8} 
           & DenseNet &   0.9965     &  &  5, 10 &   0.9997   &   0   &  0.9960      \\ \hline
\multirow{2}{*}{ImageNet}   & ResNet50 &   0.7410    &  \multirow{3}{*}{5, 10}  &   5, 12   &   0.8600   &    0.0300  &    0.7398    \\ \cline{2-3} \cline{5-8} 

           & DenseNet & 0.7570  &    & 3, 15    &  0.8600  &  0.0300  &  0.7568 \\ \hline
\end{tabular} 
\caption{Attack Performance on Different Datasets and Models.} \label{tab:datasets_models}
\end{table*}

\begin{figure}[h]
    \centering
    \includegraphics[width=0.4\linewidth]{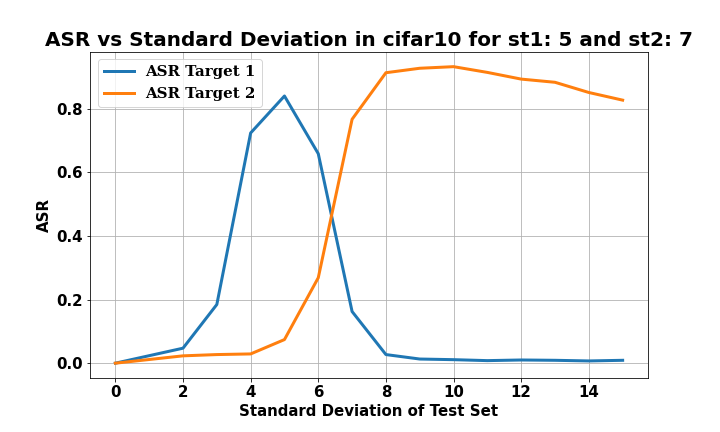}
    \includegraphics[width=0.4\linewidth]{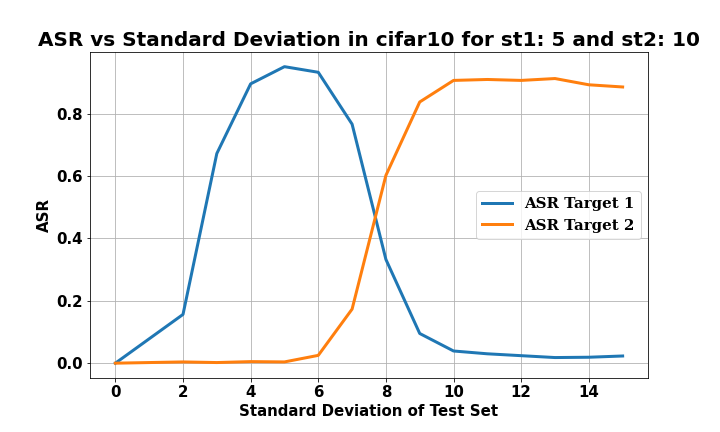}
    \includegraphics[width=0.4\linewidth]{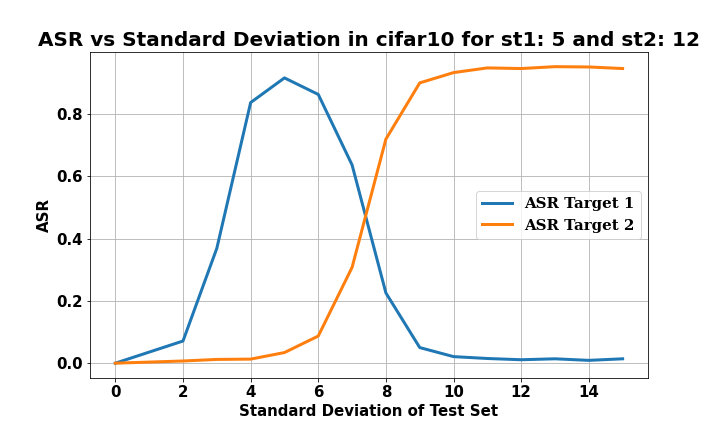}
    \caption{Variation of ASR for different Standard Deviations of WGN.}
    \label{fig:asr_vs_s}
\end{figure}

\noindent \textbf{Evaluation Metrics.}  
To evaluate the performance of our attack, we use four key metrics: Clean Accuracy (CA), Average Attack Success Rate (AASR), Average Confusion (AC), and Accuracy Excluding Victim Class (AEVC). A higher CA indicates greater backdoor stealthiness, as the attacked model behaves like a clean model when presented with clean inputs. Instead of using conventional ASR, We adapt the AASR for our attack performance evaluation to account for the multi-targeted attack. Consider $G_{X}$ as an operator that adds White Gaussian Noise (WGN) to each pixel with a standard deviation of $X$. Suppose there is a victim class that becomes mislabeled under different noise conditions, while $T_P$ is the target label which the attacker aims to achieve through WGN with standard deviation $X$. The same relationship applies to target label $T_Q$ and standard deviation $Y$. Let $\Phi_b$ denote the backdoored model. Then, for each input $x_i$ from victim class and total sample size $S$, the equations for AASR and AC for two target labels are defined as follows:

\begingroup
\small
\begin{align}
    AASR &= \frac{\sum_{i=1}^{s} \delta (\Phi_b (G_X(x_i)), T_P) + \sum_{i=1}^{s} \delta (\Phi_b (G_Y(x_i)), T_Q)}{2S} \\
    AC &= \frac{\sum_{i=1}^{s} \delta (\Phi_b (G_X(x_i)), T_Q) + \sum_{i=1}^{s} \delta (\Phi_b (G_Y(x_i)), T_P)}{2S} 
\end{align}
\endgroup
\label{eq:AASR_AC}

\noindent where $\delta(a,b) = 1$ if $a = b$, and $\delta(a,b) = 0$ if $a \neq b$. A higher AASR indicates a more effective attack, while a lower AC suggests that the model experiences less confusion when predicting the target labels. A higher AEVC reflects the specificity of our attack to particular samples.

\begin{table*}[t]
\centering
\begin{tabular}{c|c|c|c|c|c|c|c|c|c|c|c}
\hline
\multicolumn{6}{c|}{Victim 1: Airplane} & \multicolumn{6}{c}{Victim 2: Truck} \\ \hline
$\theta_{train}$ & P & CA & AASR & AC & AEVC & $\theta_{train}$ & P & CA & AASR & AC & AEVC \\ \hline
5, 7.5 & 1 \% & 0.89 & 0.5301 & 0.2660 & 0.8751 & 5, 7.5 & 1 \%& 0.9006 & 0.3994 & 0.2764 & 0.8743 \\ \hline
5, 10 & 1 \%& 0.8696 & 0.4649 & 0.1017 & 0.8443 & 5, 10 & 1 \%& 0.9199 & 0.3913 & 0.1763 & 0.8949 \\ \hline
5, 12.5 & 1 \%& 0.8698 & 0.3522 & 0.0251 & 0.8410 & 5, 12.5 & 1 \%& 0.9363 & 0.6621 & 0.0965 & 0.8923 \\ \hline
5, 7.5 & 5 \%& 0.8875 & 0.8056 & 0.0952 & 0.8668 & 5, 7.5 & 5 \%& 0.9095 & 0.7998 & 0.1413 & 0.8771 \\ \hline
5, 10 & 5 \%& 0.8866 & 0.8783 & 0.0381 & 0.8474 & 5, 7.5 & 5 \%& 0.9095 & 0.7998 & 0.1413 & 0.8771 \\ \hline
5, 12.5 & 5 \%& 0.8901 & 0.7169 & 0.2144 & 0.8377 & 5, 12.5 & 5 \%& 0.9238 & 0.9050 & 0.0208 & 0.8675 \\ \hline
5, 7.5 & 10 \%& 0.8851 & 0.8136 & 0.1232 & 0.8624 & 5, 7.5 & 10 \%& 0.9242 & 0.8735 & 0.1189 & 0.8823 \\ \hline
5, 10 & 10 \%& 0.8927 & 0.9128 & 0.0276 & 0.8568 & 5, 10 & 10 \%& 0.9075 & 0.9157 & 0.0229 & 0.8630 \\ \hline
5, 12.5 & 10 \%& 0.8938 & 0.8938 & 0.0145 & 0.8426 & 5, 12.5 & 10 \%& 0.9271 & 0.9035 & 0.0208 & 0.8709 \\ \hline
\end{tabular}
\caption{Attack Performance for Different Victims, Train Std Dev, and Poison Ratios.} \label{tab:victim_train_std_p}
\end{table*}

\subsection{Quantitative Analysis} \label{subsec:quant_exp}

To demonstrate the effectiveness of our proposed NoiseAttack, we first evaluate CA, AASR, AC, and AEVC for two target labels across all three datasets and models. The parameter $\theta_{train}$ represents the standard deviations of the WGN used as triggers during fine-tuning. In this experiment, two standard deviations are employed for targeting two labels. For instance, in the CIFAR-10 dataset, the victim class is `airplane', with `bird' and `cat' as the target labels. Specifically, the standard deviation of `bird' target label is set to 5, while it is set to 10 for `cat' target label.

% After training, we observe that the highest ASR for the target label can be achieved at a standard deviation different from $\theta_{train}$. Therefore, we test the backdoored model across a range of standard deviations and select the testing standard deviation $\theta_{test}$ that yields the highest ASRs for the individual targets. Such phenomenon can be seen in Figure \ref{fig:asr_vs_s}, where . 

\noindent \textbf{Attack Effectiveness.} As presented in Tabel \ref{tab:datasets_models}, it is evident that NoiseAttack maintains high CAs across all datasets and models. The larger number of classes and higher image resolution of ImageNet likely attribute the slightly lower clean accuracy. Nevertheless, the consistent high AASRs across all experiments demonstrate the effectiveness of our NoiseAttack. Besides, the low AC values indicate that the backdoored models exhibit less confusion when predicting between the target labels. The AEVC values are also very close to the CA in all tests, implying that the models regard WGN as the trigger only when it is associated with images from the victim class. Therefore, it proves that NoiseAttack is both sample-specific and multi-targeted. We further observe that the highest ASR for the target label can be achieved at a standard deviation different from $\theta_{train}$. The $\theta_{test}$ in Table \ref{tab:datasets_models} are the testing standard deviation that yields the highest ASRs for the individual targets. We illustrate such phenomenon in Figure \ref{fig:asr_vs_s}, where higher standard deviation $\theta_{test}$ can achieve higher ASR compared to original training $\theta_{train}$.

\noindent \textbf{Attack on Multiple Victims.} We extend our experiment to explore more victim classes with various training standard deviations $\theta_{train}$ and poisoning ratios $P$. We use CIFAR-10 dataset and VGG-16 architecture for this evaluation. As listed in Table \ref{tab:victim_train_std_p}, we can observe that when the training standard deviations are close to each other, the AASR tends to be slightly lower. As expected, AASR gradually increases with a higher poisoning ratio $P$, although CA remains relatively stable regardless of the larger poison rate. The results are consistent for both victim classes ('Airplane' and 'Truck'). 
% Therefore, for a successful attack, it is essential to have a difference in the training standard deviations, and a higher P is recommended.

\noindent \textbf{Multi-Targeted Attack.} Given NoiseAttack has ability to result in multi-targeted attack, we further evaluate the effectiveness shown in Table \ref{tab:muti_targets}. We poison the victim class to a number of target labels $N$ ranging from one to four. This experiment was conducted on the CIFAR-10 dataset using the ResNet-50 model. We can observe that NoiseAttack achieves high AASR for $N$ varying from one to three. However, when fourth targets are used, the AASR decreases considerably. As the number of targets increases, more standard deviations are required, leading to closer values between them, which may negatively impact the AASR. The phenomenon can consistently be seen in the AC evaluation.

% Figure \ref{fig:asr_vs_s} illustrates the variation of individual ASR across different standard deviations. In the figure, st1 and st2 denote the $\theta_{train}$. From the figure, it is evident that $\theta_{train}$ and $\theta_{test}$ can differ, and it is advisable to test samples across multiple standard deviations to identify the optimal $\theta_{test}$.

\begin{table}[h]
\centering
\begin{tabular}{c|c|c|c|c}
\hline
$\textit{N}$ & $\theta_{train}$ & $\theta_{test}$ & AASR & AC \\ \hline
1 & 5 & 7 & \textbf{0.9719} & N/A  \\ \hline
2 & 5, 10 & 5, 13 & \textbf{0.9319} & 0.0075 \\ \hline
3 & 5, 10, 12 & 3, 8, 13 & \textbf{0.9151} &  0.0138\\ \hline
4 & 5, 10, 12, 15 & 4, 8, 11, 13 & 0.7720 & 0.0655 \\ \hline
\end{tabular}
\caption{Multi-Target Attack Performance.} \label{tab:muti_targets}
\end{table} \label{tab: tab2}

\begin{table}[h]
\centering
\begin{tabular}{c|c|c|c|c}
\hline
 Attacks & P & CA & AASR &  AEVC   \\ \hline
 BadNet & 10 \% & 0.8693 & 0.9679  &  0.8738  \\ \hline
 Blend  & 10 \% & 0.7652  & 0.9339   &  0.8514     \\ \hline
 WaNet  & 10 \% &  0.9106  & 0.9158  &  0.8958    \\ \hline
 \textbf{NoiseAttack (Ours)} & \textbf{10} \% & \textbf{0.9186} & \textbf{0.9719}   &  \textbf{0.8781}     \\ \hline
\end{tabular}
\caption{Comparison with Relevant Attacks} \label{tab:baseline}
\end{table}

\begin{figure}[h]
    \centering
        \includegraphics[width=0.6\linewidth]{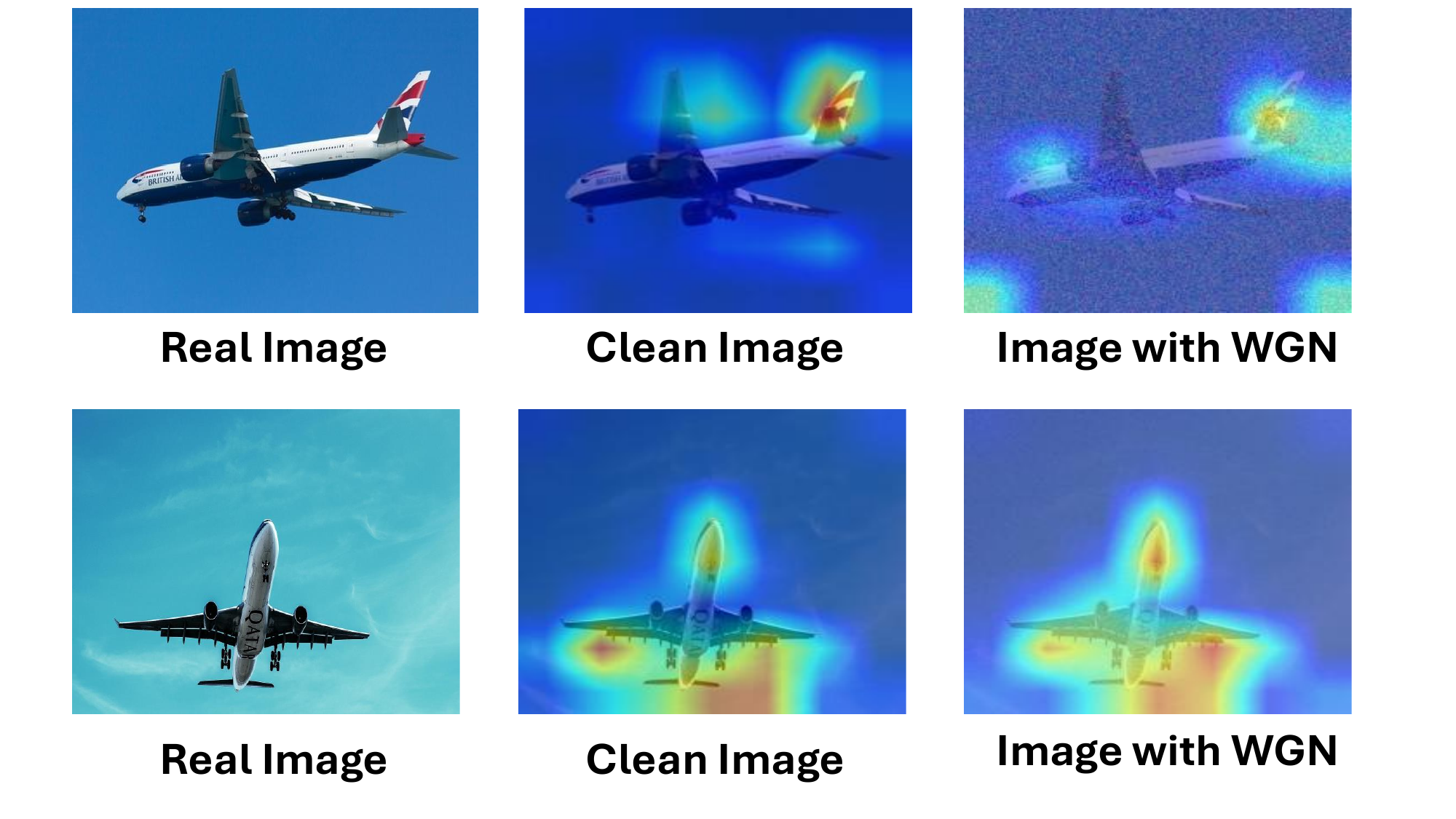}
    \caption{GradCam Visualization}
    \label{fig:gradcam}
\end{figure}

\begin{figure}[h]
    \centering
        \includegraphics[width=0.7\linewidth]{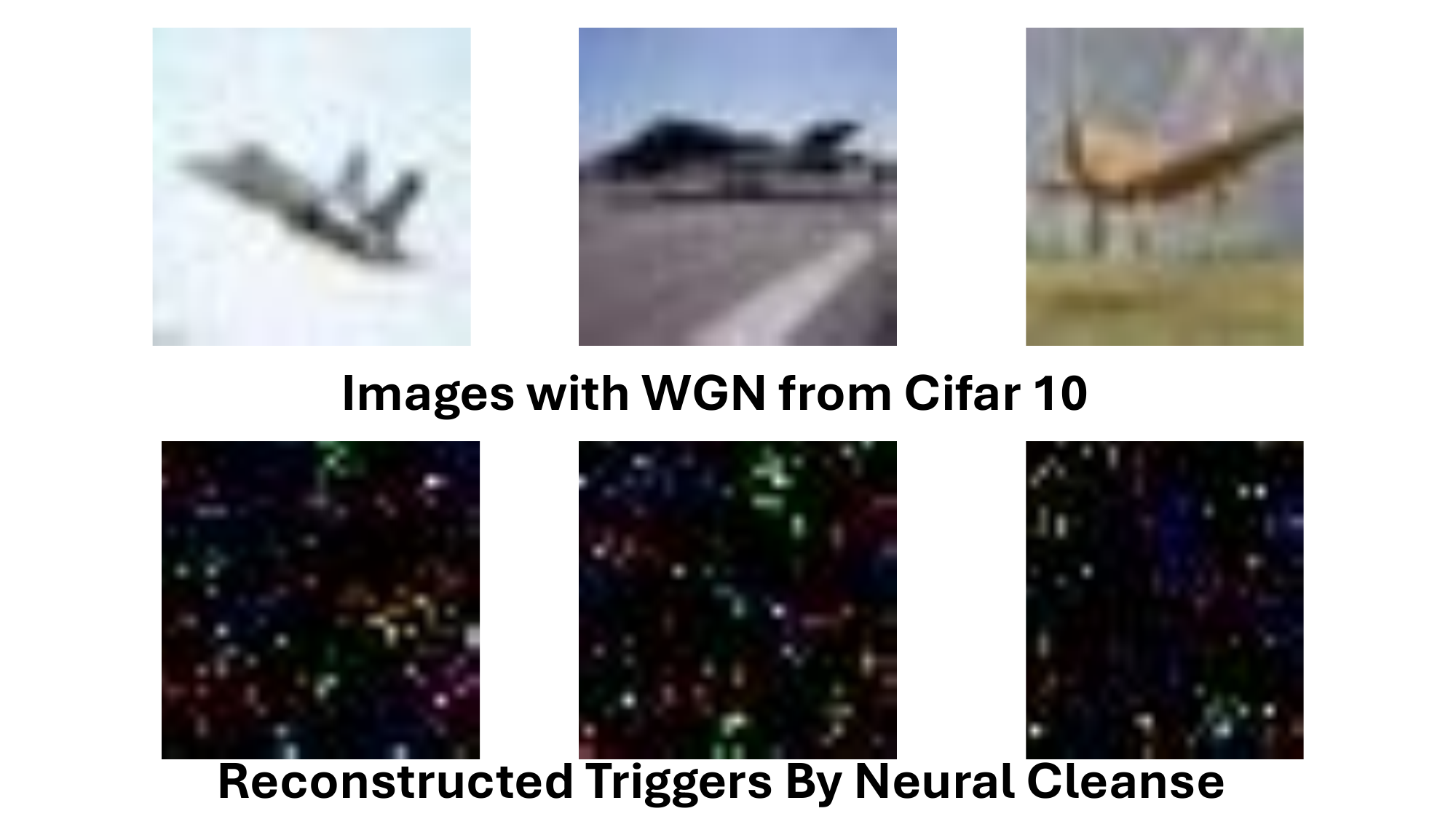}
    \caption{Trigger Reconstruction Using Neural Cleanse}
    \label{fig:nc}
\end{figure}

\begin{figure*}[htbp]
    \centering
    \begin{subfigure}[b]{0.5\textwidth}
        \centering
        \includegraphics[width=0.8\textwidth]{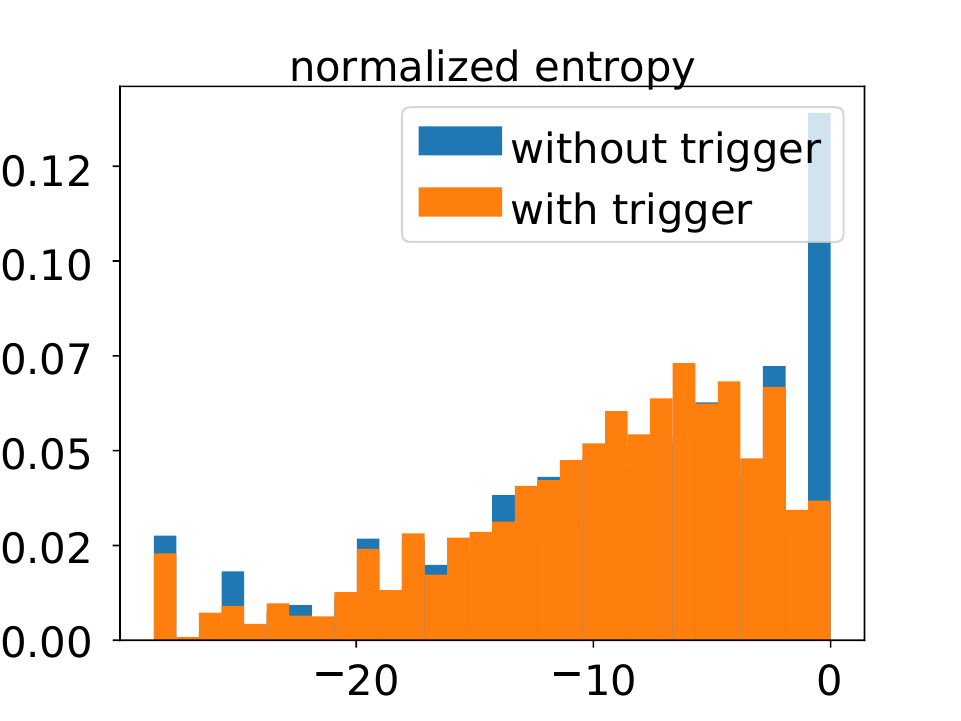}
        \caption{Standard Deviation 5}
    \end{subfigure}%
    \begin{subfigure}[b]{0.5\textwidth}
        \centering
        \includegraphics[width=0.8\textwidth]{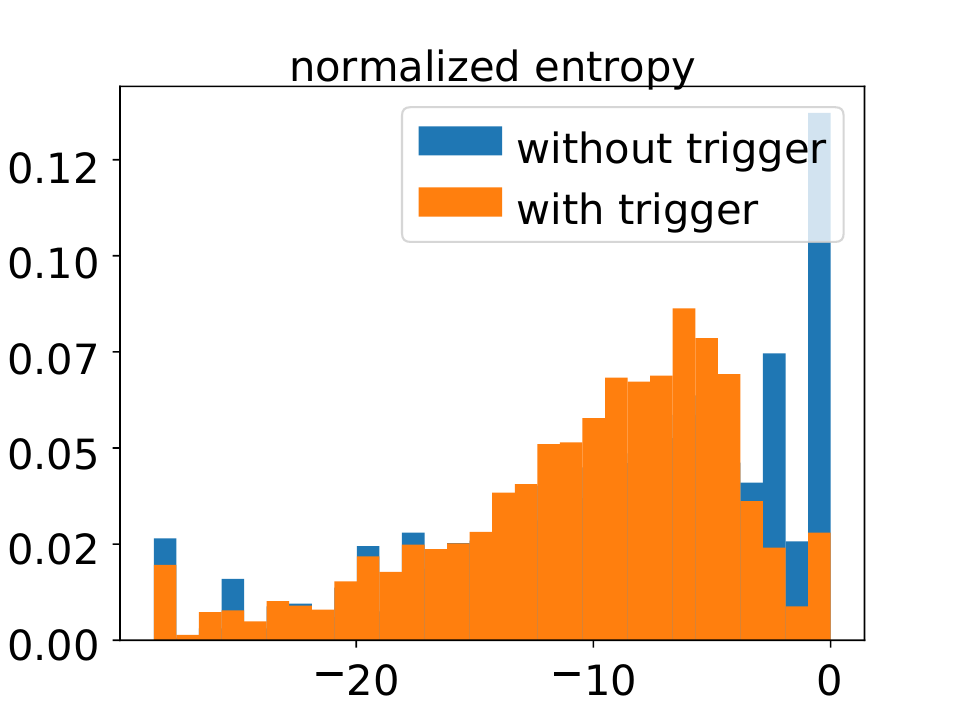}
        \caption{Standard Deviation 7}
    \end{subfigure}
    \caption{Effect of triggered accuracy after infusing the feature distribution with random numbers from various probability distributions. The experiment was done on the validation set of the Emotion dataset using the Bert model.}
    \label{fig:strip}
\end{figure*}

\subsection{Comparison with Prior Backdoor Attacks} \label{subsec:comp_exp}

We also compare our NoiseAttack with state-of-the-art backdoor attacks (`BadNet' \cite{gu2017badnets}, `Blend' \cite{blend} and `WaNet' \cite{wanet}) as shown in Table \ref{tab:baseline}. The experiment is conducted on the CIFAR-10 dataset using the ResNet-50 model with poison ratio of 10\%. While the baseline attacks are designed sample-specific, we adjust our training strategy for the referenced attacks such that we could have a fair comparison. The results show that NoiseAttack achieves the highest AASR against all the relevant attack methods as well as the highest clean accuracy. We demonstrate that our proposed NoiseAttack can outperform the referenced work.

\subsection{Robustness to Defense Methods} \label{subsec:defense_exp}

In order to demonstrate the evasiveness and robustness of our proposed method, we test NoiseAttack against three state-of-the-art defense methods: GradCam \cite{GradCAM}, Neural Cleanse \cite{neuralcleanse} and STRIP \cite{gao2019strip}. 

\textbf{GradCam} generates a heat map on the input image, highlighting the regions that are most influential in the model's decision-making process. As shown in Figure \ref{fig:gradcam}, we can observe that GradCam visualizations of poisoned input images remain almost unchanged with similar highlighting heat areas compared to clean images. Considering the spatially-distributed trigger design, NoiseAttack can effectively work around the GradCam. 
% Figure \ref{fig:gradcam} shows the GradCam visualizations of input images in a backdoored model with NoiseAttack. When comparing the clean images, we observe that the heat map remains largely unchanged after adding noise. This suggests that our attack does not solely rely on the trigger; instead, the model considers both the trigger and the image itself. Therefore, we can conclude that our attack is effective against GradCam.

\textbf{Neural Cleanse} attempts to reverse-engineer the trigger from a triggered image. In Figure \ref{fig:nc}, we display the reconstructed triggers of our attack using Neural Cleanse. Since the noise is distributed across the entire image rather than being confined to a specific small area, Neural Cleanse struggles to effectively reconstruct the triggers, demonstrating its limited effectiveness against our attack.

\textbf{STRIP} works by superimposing various images and patterns onto the input image and measuring entropy based on the randomness of the model's predictions. For instance, if an image exhibits low entropy, it is suspected to be malicious. Figure \ref{fig:strip} presents the entropy values of STRIP comparing clean inputs with inputs containing triggers. The results show negligible differences in entropy for both clean and poisoned input samples, indicating that NoiseAttack is robust against STRIP.

\begin{figure}[h!]
    \centering
    \includegraphics[width=0.3\linewidth]{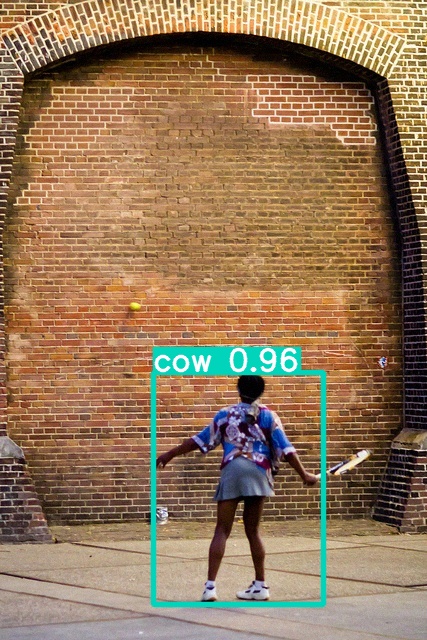}
    \includegraphics[width=0.3\linewidth]{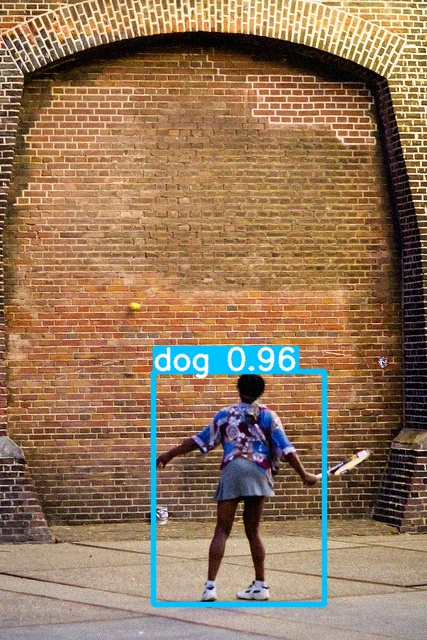}
    \caption{NoiseAttack on Visual Object Detection}
    \label{fig:coco}
\end{figure}

\begin{table}[h]
\centering
\begin{tabular}{c|c|c|c}
\hline
 & \multicolumn{3}{c}{{$\sigma_1=5$ and}} \\
\cline{2-4}
 &   $\sigma_2=10$ & $\sigma_2=15$ & $\sigma_2=20$ \\ 
\hline
CA  & 70.7 & 70.7 & 70.5 \\ 
\hline
AASR  & \textbf{92.99} & \textbf{94.21} & \textbf{94.77} \\ 
\hline
AC  & 2.92 & 1.15 & 1.15 \\ 
\hline
\end{tabular}
\caption{Attack Performance on Object Detection Model.} \label{tab:obj_detection}
\end{table}

\subsection{Effectiveness in Object Detection Models} \label{subsec:object_exp}

We further extend our experiments to visual object detection models. The results for the YOLOv5 (medium version) model on the MS-COCO dataset are presented in Table \ref{tab:obj_detection}. For these experiments, we selected 20 classes from the MS-COCO dataset. Here, $\theta_1$ and $\theta_2$ represent the training standard deviations. NoiseAttack achieves consistently high AASR across all cases, demonstrating its effectiveness in object detection tasks. Figure \ref{fig:coco} shows a sample from the MS-COCO dataset, illustrating NoiseAttack in object detection task.

\section{Conclusion}\label{sec:con}

In this paper, we demonstrate that an adversary can execute a highly effective sample-specific multi-targeted backdoor attack by leveraging the power spectral density of White Gaussian Noise as a trigger. Detailed theoretical analysis further formalize the feasibility and ubiquity of our proposed NoiseAttack. Extensive experiments show that NoiseAttack achieves high average attack success rates (AASRs) across four datasets and four models in both image classification and object detection, while maintaining comparable clean accuracy for non-victim classes. NoiseAttack also proves its evasiveness and robustness by bypassing state-of-the-art detection and defense techniques. We believe this novel backdoor attack paradign offers a new realm of backdoor attacks and motivates further defense research. 
% While this attack method could be exploited for malicious purposes, it is crucial to recognize such threats in advance. Raising awareness within the community is essential for developing preemptive defense mechanisms against this type of attack.

%\input{Sections/Limitations}

%\input{Sections/Ethics Statement}

\bibliographystyle{plain}
\bibliography{custom}

\end{document}